\documentclass[10pt,twocolumn,letterpaper]{article}

\usepackage{iccv}
\usepackage{times}
\usepackage{epsfig}
\usepackage{graphicx}
\usepackage{amsmath}
\usepackage{amssymb}

\usepackage{caption}

\usepackage{textcomp, gensymb}
\usepackage{multirow}
\usepackage{booktabs}
\usepackage{pifont} 
\usepackage[dvipsnames]{xcolor}
\usepackage{tabularx}
\usepackage{enumitem}
\usepackage{color, colortbl}
\usepackage[accsupp]{axessibility}  

\newcommand{\cmark}{\textcolor{ForestGreen}{\ding{51}}}
\newcommand{\xmark}{\textcolor{red}{\ding{55}}}
\newcommand{\gazevector}{$\mathbf{v}_g \,$}

\newcommand{\gazedecoder}{$\mathcal{GOT} \,$}

\usepackage[pagebackref=true,breaklinks=true,letterpaper=true,colorlinks,bookmarks=false]{hyperref}

\iccvfinalcopy 

\def\iccvPaperID{9004} 

\ificcvfinal\fi

\begin{document}

\title{Object-aware Gaze Target Detection}

\author{Francesco Tonini$^{1,2}$, Nicola Dall'Asen$^{1,3}$, Cigdem Beyan$^{1}$, Elisa Ricci$^{1,2}$\\
{\normalsize\centering$^{1}$ University of Trento, Trento, Italy}~
{\normalsize\centering$^{2}$ Fondazione Bruno Kessler, Trento, Italy}\\
{\normalsize\centering$^{3}$ University of Pisa, Pisa, Italy}\\
{\tt\small \{francesco.tonini, nicola.dallasen, cigdem.beyan, e.ricci\}@unitn.it}
}

\maketitle

\ificcvfinal\thispagestyle{empty}\fi

\begin{abstract}
Gaze target detection aims to predict the image location where the person is looking and the probability that a gaze is out of the scene.
Several works have tackled this task by regressing a gaze heatmap centered on the gaze location, however, they overlooked decoding the relationship between the people and the gazed objects. This paper proposes a Transformer-based architecture that automatically detects objects (including heads) in the scene to build associations between every head and the gazed-head/object, resulting in a comprehensive, explainable gaze analysis composed of: gaze target area, gaze pixel point, the class and the image location of the gazed-object. Upon evaluation of the in-the-wild benchmarks, our method achieves state-of-the-art results on all metrics (up to 2.91\% gain in AUC, 50\% reduction in gaze distance, and 9\% gain in out-of-frame average precision) for gaze target detection and 11-13\% improvement in average precision for the classification and the localization of the gazed-objects.
The code of the proposed method is publicly available\footnote{\url{https://github.com/francescotonini/object-aware-gaze-target-detection}}.
\end{abstract}

\vspace{-0.35cm}
\section{Introduction}
\label{sec:intro}
\vspace{-0.2cm}
Gazing is a powerful nonverbal signal, which indicates the visual attention of a person and allows one to understand the interest, intention, or (future) action of people \cite{emery2000eyes}. For this reason, gaze analysis has widely been used in several disciplines such as human-computer interaction \cite{noureddin2005non,yin2014real}, neuroscience \cite{dalton2005gaze,rayner1998eye}, social and organizational psychology \cite{capozzi2019tracking,edwards2015social}, and social robotics \cite{admoni2017social} to name a few. 

Even though human beings have a remarkable capability to decode the gaze behavior of others, realizing this task \emph{automatically} remains a challenging problem \cite{bao2022escnet,tonini2022multimodal,tu2022end}. The computer vision community has tackled the automated gaze behavior analysis in terms of two tasks: (a) \emph{gaze estimation} and (b) \emph{gaze target detection}. Gaze estimation stands for predicting the person's gaze direction (usually in 3D) when typically a \emph{cropped human head image} is given as the input \cite{Cheng_2018,funes2014eyediap,guo2020domain,kellnhofer2019gaze360}. Instead, gaze target detection (also referred to as gaze-following) is to determine the specific (2D or 3D) location that a human is looking at in an \emph{in-the-wild} scene~\cite{chong2020detecting,fang2021dual,Zhengxi2022}.

\begin{figure}[t!]
\centering
\resizebox{1.\linewidth}{!}{%
\begin{tabular}{rl}
\raisebox{0.20\height }{\rotatebox[origin=l]{90}{Traditional}} & \includegraphics[width=0.9\linewidth]{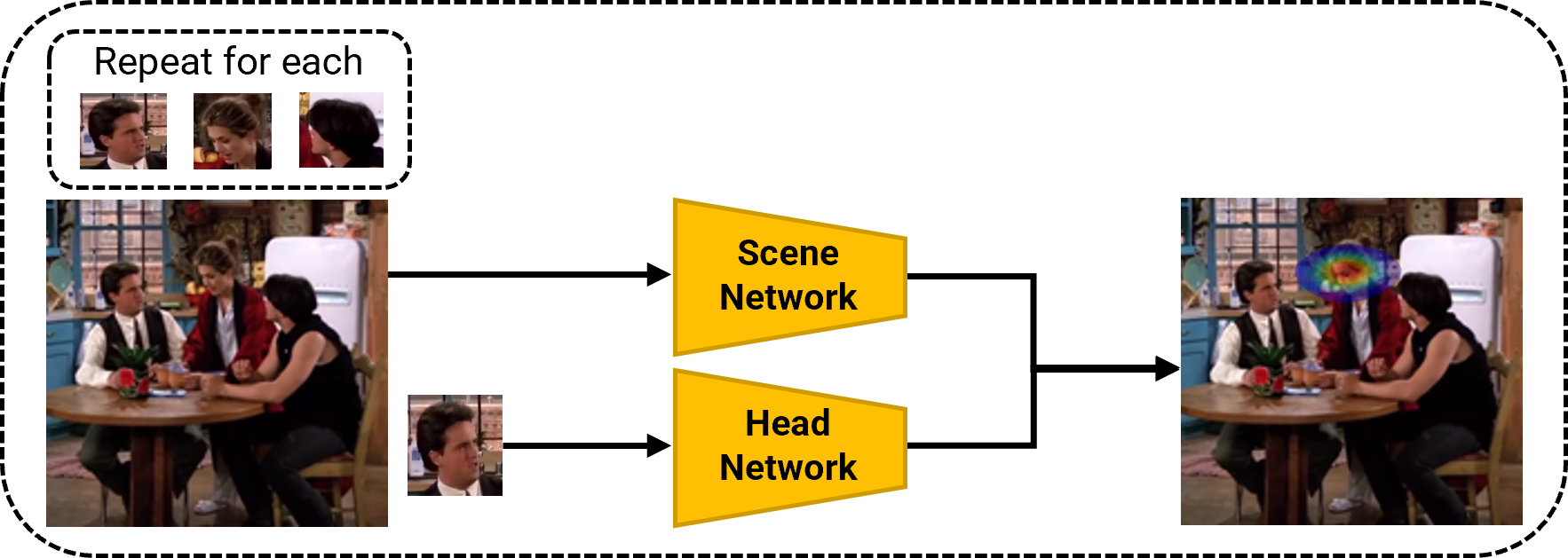}  \\ 
\raisebox{-0.07\height }{\rotatebox[origin=l]{90}{Tu et al.~\cite{tu2022end}}} & \includegraphics[width=0.9\linewidth]{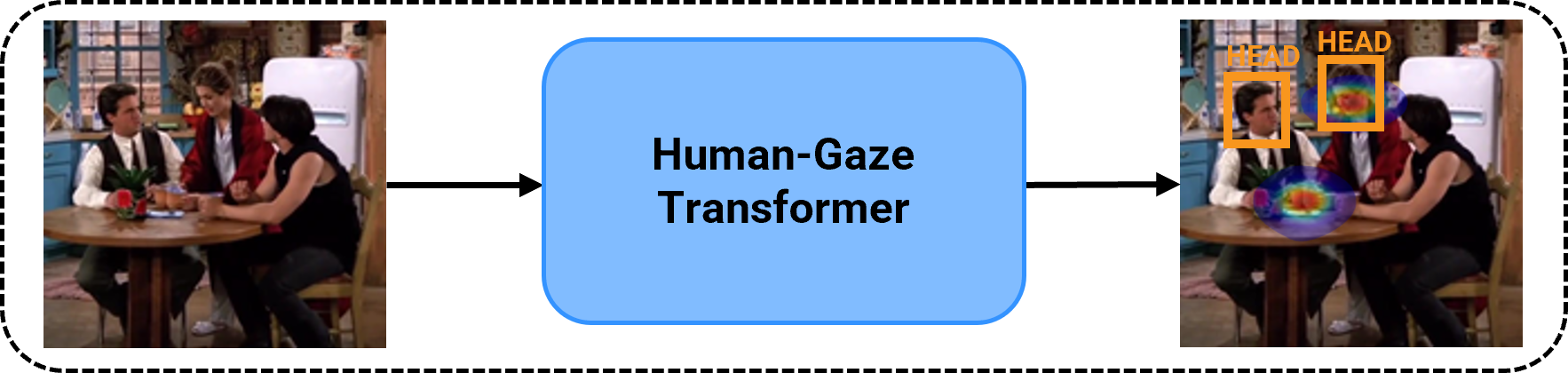} \\ 
\raisebox{0.60\height }{\rotatebox[origin=l]{90}{Ours}} &  \includegraphics[width=0.9\linewidth]{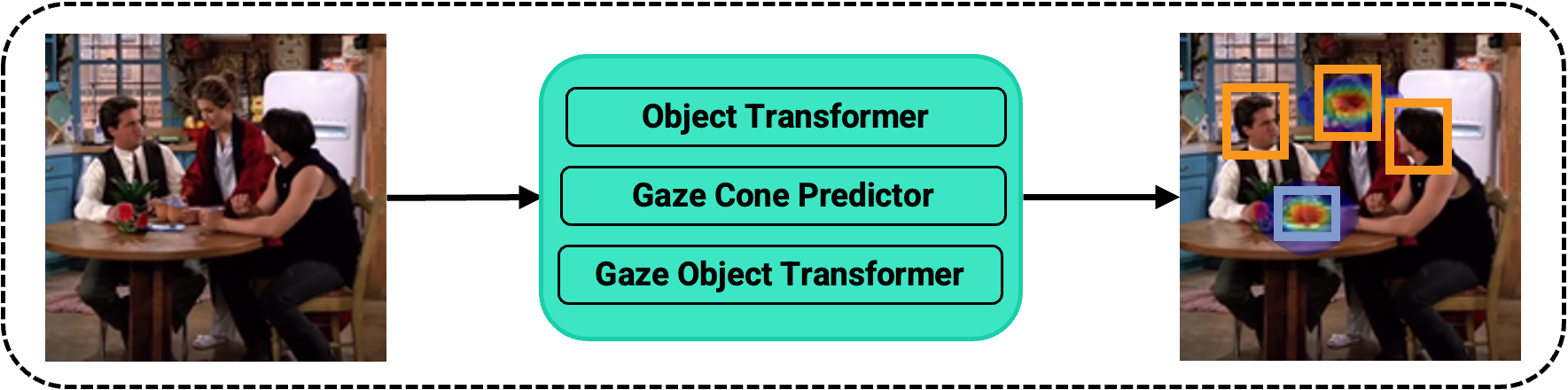}\\ 
\end{tabular}
}
\vspace{-0.3cm}
\caption{The overall methodology of the existing approaches and ours.}
\label{fig:teaser}
\vspace{-0.4cm}
\end{figure}

Several works utilize head pose features and the saliency maps of possible gaze targets to perform gaze target detection.
For instance,~\cite{Chong_2018_ECCV,lian2018believe,Recasens2015,Recasens2017} followed a two-pathway learning scheme, where one path learns feature embeddings from the scene image, and the other path models the head crops belonging to the person whose gaze target is aimed to be predicted. Chong \textit{et al.}~\cite{chong2020detecting} extend the aforementioned two-pathway approach to perform spatio-temporal modeling to determine the gaze targets in videos. In the same vein, a few other methods exist: \cite{bao2022escnet,fang2021dual,jin2022depth,miao2023patch,tonini2022multimodal}. Among them, some consider a third path to model the depth map of the scene image, which is determined by a monocular depth estimator
\cite{fang2021dual,jin2022depth,miao2023patch,tonini2022multimodal}. Differently, others \cite{bao2022escnet} inject depth maps and 2D-human poses to improve the 3D understanding of the scenes, resulting in better gaze target detection. 
The results achieved by these approaches (referred to as \emph{traditional methods} throughout the manuscript, see Fig.~\ref{fig:teaser}-top) are highly remarkable since they demonstrated that gaze target estimation could be directly performed on images or videos in contrast to using low-intrusive wearable eye trackers, which notoriously have several issues in terms of cost, battery life, and calibration. On the other hand, traditional methods also have some major drawbacks.  First, both training and inference require carefully human-annotated head crops. Therefore, to ensure that traditional methods work well in real-life practical applications, there is a need for additional and highly accurate head detectors. Indeed, Tu \textit{\textit{et al.}} \cite{tu2022end} showed notable performance drops of traditional methods when head detectors were involved instead of using manually annotated head locations. A second limitation concerns the fact that traditional methods can perform a single gaze target detection at a time; thus, for scenes containing multiple people, the models should be run repeatedly for each person. Besides the computational complexity such implementation brings in, post-processing is also needed to combine the detected gaze targets of different subjects in the same scene. Tu \textit{et al.} \cite{tu2022end} to some extent overcome the shortcomings mentioned above by introducing a Transformer-based architecture that explicitly learns how to detect and localize the head during gaze target detection (see Fig.~\ref{fig:teaser}-middle). 
However, the contribution of objects to decipher the human-human/object gazing is completely omitted in \cite{tu2022end}.

Several studies show that people typically gaze at living or non-living \emph{objects} in the scene during social and physical interactions \cite{cho2021human,li2021eye,mazzamuto2022weakly,Schauerte2014,10023877,yuguchi2019real,yun2013exploring}. Motivated by this, we pursue an object-aware gaze target detection, instead of using features extracted from holistic scene images and head crops as in traditional methods: ~\cite{bao2022escnet,Chong_2018_ECCV,chong2020detecting,fang2021dual,jin2022depth,lian2018believe,miao2023patch,Recasens2015,tonini2022multimodal} or learning how to detect and localize the head of the person-in-interest (the one whose gaze target to be detected) as in \cite{tu2022end}. Our proposal is not only able to predict the \emph{gaze area} (in terms of heatmaps) that people looking at and determine if the gaze target is inside or outside of the scene but also \emph{localize the objects} and \emph{predicts the objects' classes} (including head) on which the gaze point is (see Fig.~\ref{fig:teaser}-bottom). The further has significant practical usage since it brings in an \emph{explainable} gaze analysis (see Table~\ref{tab:previous} for details).

The proposed method is an end-to-end Transformer-based architecture. Given a scene image, we first extract all objects, including the ones classified as heads, with an \emph{Object Detector Transformer}. Then, for each head, a gaze vector is predicted. Using this gaze vector, we build a \emph{gaze cone} for each person individually, allowing the model to filter out the objects that are not in a person's Field of View (FoV). Subsequently, a masked transformer (called \emph{Gaze Object Transformer}) learns the interactions between the detected heads and objects, boosting the gaze target detection performance in terms of both heatmaps and gaze points (\ie a single pixel in the scene). Furthermore, this architecture has a remarkable capability to predict whether a gaze target point is out of the frame. The extensive evaluations on two large-scale benchmark datasets show the superior performance of our method \textit{w.r.t.} state-of-the-art (SOTA) gaze target detectors. At the same time, our model has additional competence to accurately predict the gazed objects' locations and the associated classes as empirically demonstrated. The ablation study highlights the importance of all components and specifically the needs for our main technical contributions, \ie the Gaze Cone Predictor and the Gaze Object Transformer.

To summarize: (1) We introduce a novel object-oriented gaze target detection method. (2) This end-to-end Transformer-based model automatically detects the heads and other objects in the scene to build associations between every head and the gazed-head/object, resulting in a comprehensive, explainable gaze analysis composed of: gaze target area, gaze pixel point, the class of gazed-object, the bounding box of the gazed-object as well as predicting whether the gazed point is out of the frame. (3) We demonstrate SOTA results on standard datasets regarding all evaluation metrics for gaze-target detection (up to 2.91\% gain in AUC, 50\% reduction in gaze distance, and 9\% gain in out-of-frame AP), gazed-object classification and localization (11-13\% gain in AP) and in case of low/high variance across gaze annotations. (4) The code of the proposed method is publicly available. We also release our implementation\footnote{\url{https://github.com/francescotonini/human-gaze-target-detection-transformer}} for \cite{tu2022end} since during our private communications with the authors, we are informed that their code at the moment cannot be shared by them due to their ongoing collaborations with a company.

\begin{table}[!tb]
\centering
\resizebox{0.95\linewidth}{!}{
\begin{tabular}{lccccc}
\hline
\multicolumn{1}{c}{\multirow{2}{*}{Method}} & Wout/ Head   & Multiple & Head      & Object       & Object         \\
\multicolumn{1}{c}{}                        & Loc. Given & People   & Detection & Localization & Classification \\ \hline
Traditional                                 & \xmark     & \xmark   & \xmark    & \xmark       & \xmark         \\
Tu \textit{et al.} \cite{tu2022end}         & \cmark     & \cmark   & \cmark    & \xmark       & \xmark         \\
Ours                                        & \cmark     & \cmark   & \cmark    & \cmark       & \cmark         \\ \hline
\end{tabular}}
\vspace{-0.2cm}
\caption{Existing gaze target detection methods compared to ours. Ours is more explainable since, for every person in the scene, it can detect the object class and bounding box location on which the gaze is. It learns the scene objects (including the head) without using the head locations supplied by datasets.}
\vspace{-0.6cm}
\label{tab:previous}
\end{table}

\begin{figure*}[!th]
\begin{center}
\includegraphics[width=.75\textwidth]{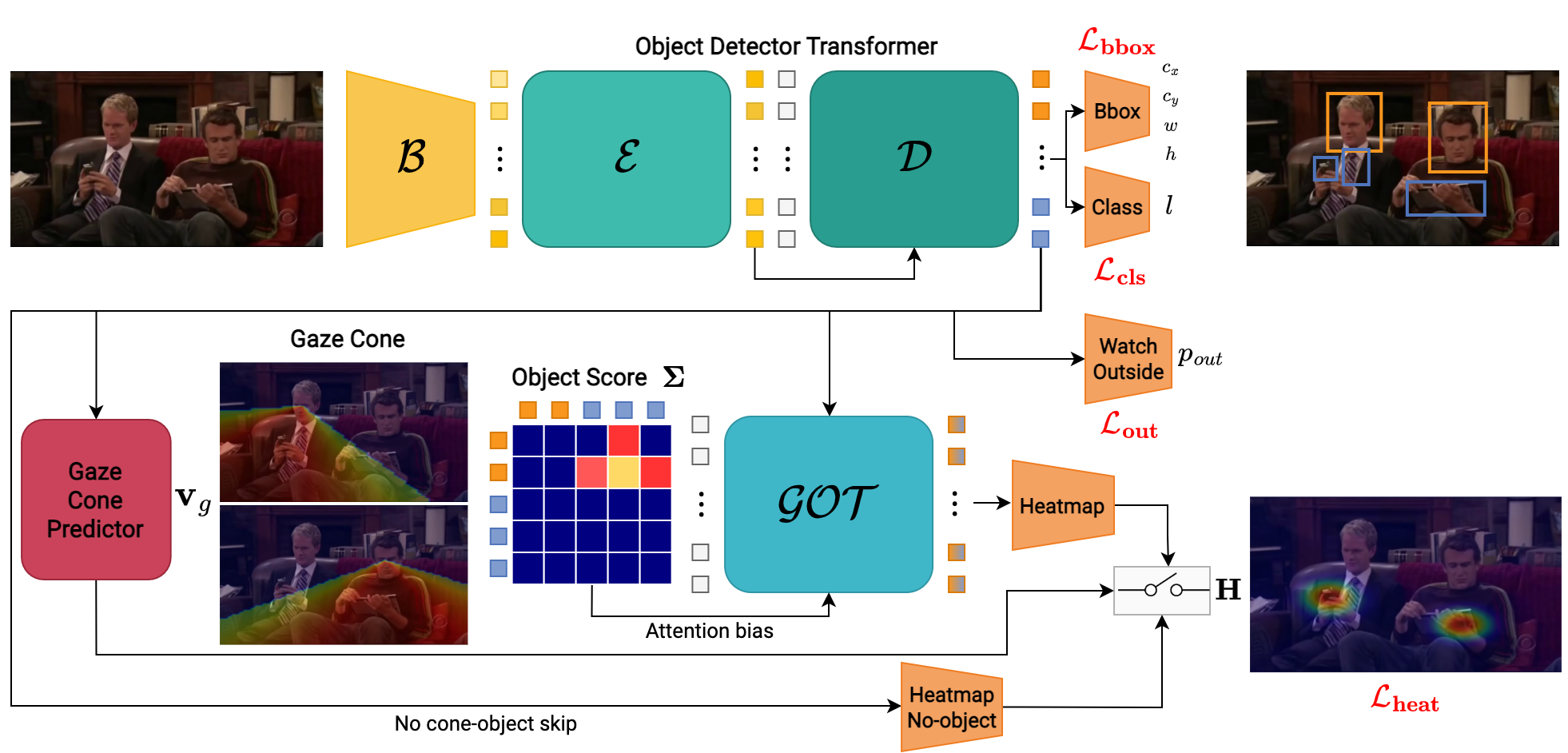}
\end{center}
\vspace{-0.5cm}
   \caption{\textbf{Proposed method.} The encoder ($\mathcal{E}$) and decoder ($\mathcal{D}$) of the Object Detector Transformer operate on the features extracted by a backbone $\mathcal{B}$ to learn rich object features used to detect and localize objects (including heads) in the scene. Head features are used to build the \textbf{gaze cone}. Objects in the cone are extremely likely to be gaze-interesting. The \textit{object score} matrix $\Sigma$ boosts attention scores in the \textit{Gaze Object Transformer} (\gazedecoder), whose output features are used to build the gaze heatmap. If no object lies in the cone, a skip-connection lets the network predict the heatmap from head features only.}
\label{fig:architecture}
\vspace{-0.5cm}
\end{figure*}

\section{Related Work}
\label{sec:related}
\vspace{-0.2cm}
The main focus of this paper is to determine the location that a human is looking at in an in-the-wild scene captured from the third-person view.
To this end, Recasens \textit{et al.} \cite{Recasens2015} presented the first relevant dataset called GazeFollow, and proposed a two-branch Convolutional Neural Network (CNN) whose first branch estimates the saliency from scene images and the second branch processes manually annotated head crops together with their location information. Several subsequent works~\cite{bao2022escnet,chong2020detecting,fang2021dual,jin2022depth,lian2018believe,Recasens2017,tonini2022multimodal} adopted this two-branch structure and further introduced additional components. For instance, Chong \textit{et al.}~\cite{Chong_2018_ECCV} brought in detecting the gaze targets not-being in the scene (so-called \emph{out-of-frame} detection). Later on, the same authors~\cite{chong2020detecting} integrated a CNN-LSTM into their pipeline, modeling the dynamics of gaze in \emph{videos} and making frame-based inferences. They also introduced the VideoAttentionTarget dataset, made of videos.
A few studies incorporated the depth maps obtained from monocular depth estimators in addition to the embeddings learned from RGB scenes and head crops~\cite{fang2021dual,jin2022depth,miao2023patch,tonini2022multimodal}. Fang \textit{et al.}~\cite{fang2021dual} integrated the precise detection of head pose and the location of eyes. Jin \textit{et al.}~\cite{jin2022depth} used two auxiliary networks, one to learn depth features and the other to compute 3D-gaze orientation features. However, detecting head pose, eye locations, etc., are already challenging tasks to perform in-the-wild, and their inaccurate results can affect gaze target detection negatively. A better solution could be leveraging the collaborative learning of scene, depth, and head features, as shown in~\cite{tonini2022multimodal}.
Bao \textit{et al.}~\cite{bao2022escnet} proposed a method taking an RGB image and a head crop at a time and further using the depth map and a 2D human body pose detector to reconstruct the 3D scene with point clouds. Such a model~\cite{bao2022escnet} requires several detectors to be fine-tuned and therefore increases the computational complexity. Furthermore, it underperforms compared to, e.g.,~\cite{fang2021dual,tonini2022multimodal} on the VideoAttentionTarget dataset. Qiaomu \textit{et al.}~\cite{miao2023patch} used the same modalities as~\cite{tonini2022multimodal} but also included a temporal attention model and replaced the in/out prediction encoder of~\cite{chong2020detecting} with a patch distribution prediction module, resulting in effective performance in case of large annotation variances. Unlike aforesaid approaches, aka \emph{traditional methods}, using pretrained CNN backbones, Tu \textit{et al.}~\cite{tu2022end} introduced the first Transformer-based approach, outperforming the others.
Drastic performance drops for traditional methods were also demonstrated in \cite{tu2022end}, when they were evaluated with the head locations predicted by automated head detectors.

We stand out from the prior art as our method performs simultaneous gaze target detection of multiple persons in the scene by mutually learning localization and classification of the gazed-objects (including the head) and determining the head-head/object gaze interactions. Our end-to-end Transformer-based model explicitly aims to provide explainable gaze target detection, which has not been accomplished before (see Table \ref{tab:previous} for comparisons).

\section{Method}
\label{sec:method}
\vspace{-0.2cm}
The proposed method is shown in Fig~\ref{fig:architecture}. Given an image, we first predict the set of objects $\mathbf{O}=\{ \, (c_x, c_y, w, h, l)\}$ in it, where $(c_x, c_y, w, h)$ represent the center coordinates of a single object and its width and height, respectively, $l \in [0, CLS)$ is an object's label, and $CLS$ is the number of classes, including a special \textit{no object} ($\emptyset$) class (described in Sec.~\ref{sec:impDet}). To this end, after extracting the image features through a backbone $\mathcal{B}$, we use an \textit{Object Detector Transformer} that reasons on the scene features with the encoder $\mathcal{E}$ and learns relevant object features with the decoder $\mathcal{D}$. Such features are used to differentiate between heads $\mathbf{O_h}$ and other objects in the scene. For each head $\mathbf{O}_h^i$, we feed its features to the \textit{Gaze Cone Predictor} to determine a gaze vector $\mathbf{v}_g^i$ that represents the gaze direction of the person. This gaze vector is used to build a gaze cone with an angle of $\alpha$ corresponding to the Field of View (FoV) and to selectively maintain the objects that are inside the cone for each head. The \textit{Gaze-Object Transformer} ($\mathcal{GOT}$) models the relationships between the detected
objects and predicts the probability of them being the gaze target of any person, with a higher probability for the objects closer to the gaze vector. The gaze of each person is represented as a Gaussian heatmap $\mathbf{H}^i$ centered on the gaze point $\mathbf{p}_g^i$, and when no object is present inside the gaze cone, we use a \textit{no cone-object skip} to compute the heatmap directly from the head features. We also use the head features to predict the probability of the gaze target being outside the frame.
To sum up, our model consists of three major components: (a) Object Detector Transformer, (b) Gaze Cone Predictor, and (c) Gaze Object Transformer, which are described thoroughly in the following sections.

\subsection{Object Detector Transformer}
\vspace{-0.1cm}
Given an RGB image $\mathbf{x} \in \mathbb{R}^{C \times H \times W}$, we aim to predict the bounding boxes and labels of objects. We start by extracting a feature map $\mathbf{f_b} \in \mathbb{R}^{C_b \times H_b \times W_b}$ with a convolutional backbone $\mathcal{B}$, and we linearly project the channel dimension to a lower space $C^{b'}$ due to the high channel dimensionality. 
We flatten the spatial dimensions and obtain $\mathbf{f_b^{'}} \in \mathbb{R}^{H_bW_b\times C_b^{'} }$, which is fed to a transformer encoder $\mathcal{E}$ that enhances the coarse image features extracted by $\mathcal{B}$.

$\mathcal{E}$ is designed as a stack of multi-head self-attention (MHSA) and feed-forward (FFN) layers.
The projected output of $\mathcal{B}$, $\mathbf{f_b^{'}}$, forms the input queries $Q$, keys $K$, and values $V$ of $\mathcal{E}$. To retain the spatial information of the feature map, we add positional encodings for $Q$ and $K$.
The output of the encoder, $\mathbf{f_e}$, forms the input $K$ and $V$ of the cross-attention module of the transformer decoder $\mathcal{D}$.

$\mathcal{D}$ completes our Object Detector Transformer and introduces a multi-head cross-attention module to obtain object-relevant features.
First, the decoder performs self-attention on a set of learnable embeddings $\mathbf{e_d} \in \mathbb{R}^{N \times C_b^{'}}$, where $N$ is the maximum number of objects to be predicted.
Similar to $\mathcal{E}$, we add the learnable embeddings $\mathbf{e_d}$ with a set of fixed positional embeddings.
The output of the self-attention on $\mathbf{e_d}$ is then fed to a multi-head cross-attention module, where $\mathbf{e_d}$ are the queries, and $\mathbf{f_e}$ are the keys and values.
The output features $\mathbf{f}_d$ of the transformer decoder are finally used by two multi-layer perceptrons (MLP) to predict the object bounding box (Bbox) and class, respectively.

\begin{figure}[t!]
\begin{center}
\includegraphics[width=0.8\linewidth]{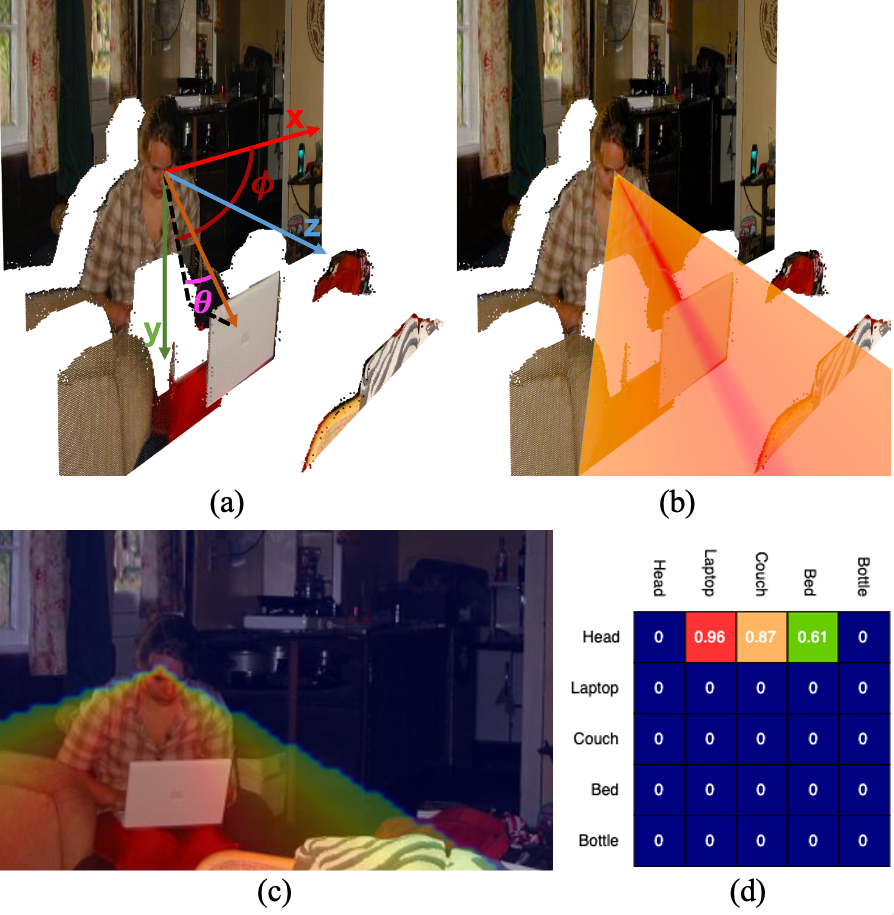}
\end{center}
\vspace{-0.7cm}
   \caption{(a) Our 3D cone construction from the head center point; $\theta$ and $\phi$ are the polar and azimuthal angle, respectively. Exploded view computed with the depth map. (b) 3D gaze cone considers the depth and, in this way, excludes the objects unrelated to the gaze vector, in this case, the couch and the bed on the right. Instead, in (c) 2D gaze cone, the couch is inside it, although common sense would tell that it should not be. (d) \textit{Object score} matrix $\Sigma$ of (c) when the detected objects are: \textit{head}, \textit{laptop}, \textit{couch}, \textit{bed}, and \textit{bottle} classes.}
\label{fig:cone}
\vspace{-0.5cm}
\end{figure}

\subsection{Gaze Cone Predictor}
\vspace{-0.1cm}
The objects predicted by the \textit{Object Detector Transformer} may appear in an area outside the FoV of a person but inside of another person(s).
Since our method performs \emph{multi-person} gaze prediction, we must consider the mentioned scenario and selectively focus on objects in the FoV of each individual.
To this end, the \textit{Gaze Cone Predictor} produces a gaze cone for each head detected and allows the \gazedecoder to focus on only the relevant objects on a person-by-person basis. For the objects detected as a head, the gaze cone, which can be either in 2D or 3D, is built based on the estimated gaze vector. The gaze cone allows us to build an \textit{object score} matrix ($\Sigma$) based on the relationship between head-head/objects.

In detail, an MLP takes as input the features of objects detected as heads $\mathbf{O}_h$ and estimates, for each of them, a 3D gaze vector $\mathbf{v}_g^i$ $= (\theta^i, \phi^i, \, \rho^i)$. Each gaze vector uniquely identifies the orientation of the person's gaze with $\theta$, $\phi$, and $\rho$, which are the polar angle, azimuthal angle, and magnitude of the vector, respectively.
For each gaze vector $\mathbf{v}_{g}^i$, we design a 3D cone of angle $\alpha$ and apex ($c_x^i, \, c_y^i, \, c_z^i $) representing the FoV of a person, where $c_x^i$, $c_y^i$, and $c_z^i$ are the center coordinates of the head.
The cone axis has the same direction as the gaze vector, and the intensity of the cone, \ie, the point saliency, is calculated as the cosine similarity between $\mathbf{v}_{g}^i$ and all vectors inside the cone starting from ($c_x^i, \, c_y^i, \, c_z^i$). In the 2D case, $\theta$ is not available, and we only have one angle $\phi$ and the magnitude $\rho$ for the gaze vector, while the 2D cone is still in the center of the apex but spans only in 2D instead of 3D. We adopt the discretized space of the same dimensionality of the predicted heatmap presented in \cite{gupta2022modular}, while we extend it to the 3D case, with $x$, $y$, and $z$ axis corresponding to the width, height, and depth of the image.
For the 2D cone building, we follow the approach of \cite{gupta2022modular}, but we constrain the cone to be a fixed angle $\alpha$, which is in line with the FoV of human boundaries \cite{howard1995binocular}. 

Below, the derivations are given for the 3D gaze cone, but the corresponding 2D implementation of them is the same except for not having the depth coordinates as described above. Refer to the visual explanation of the 2D and 3D gaze cone in Fig.~\ref{fig:cone}a-c. 
Formally, let $angle(\textbf{v}_a, \textbf{v}_b)$ be the absolute value of the angle between two vectors, and $\sigma(\mathbf{v}_a, \mathbf{v}_b)$ be the cosine similarity between two vectors $\mathbf{v}_a$ and $\mathbf{v}_b$ conditioned on the cone angle $\alpha$:
\vspace{-0.2cm}
\begin{equation}
\sigma(\mathbf{v}_a, \mathbf{v}_b)=
\begin{cases}
\cos{(\mathbf{v}_a, \mathbf{v}_b)} & \text{if} \,\, angle(\mathbf{v}_a, \mathbf{v}_b) \le \frac{\alpha}{2},  \\
0 & \text{otherwise}
\end{cases}
\vspace{-0.2cm}
\end{equation}
The projected 3D gaze cone of a person $i$, $\mathbf{CD}_{3D}^i$, whose head center coordinates are $c_x^i, c_y^i, c_z^i$, and predicted gaze vector $\mathbf{v}_g^i$, is defined as:
\vspace{-0.2cm}
\begin{equation}
    \begin{aligned}
       \small{ \mathbf{CD}_{3D}^i = \{ \sigma(\mathbf{v}_{g}^i, \mathbf{v}_{H}^{ijkl}) \} } \\ 
        \,
      \small {  \forall j, k, l \in [0, w) \times [0, h) \times [0, d) }
  \end{aligned}
 \vspace{-0.1cm}
\end{equation}
where $w$, $h$, and $d$ are the width, height, and depth of the space on which the 3D cone is computed, and $\mathbf{v}^i_H$ indicates the vectors in the discretized space starting from $(c_x^i, c_y^i, c_z^i)$.

The set of 3D cones $\mathbf{CD}_{3D}$ allows us to define the \textit{object score} as a square matrix $\Sigma$ of size $N \times N$, where $N$ is the number of objects detected by the Object Detector Transformer.
The object score matrix represents whether an object is in the visual cone of each person and how close it is to their predicted gaze vector (see Fig.~\ref{fig:cone}d).
Each row represents an object where the rows of objects not classified as heads are zero.
For rows of \textit{head} objects, the score for each other object is equivalent to the value of the gaze cone picked at the center coordinates of the object.
When no object is in the gaze cone, the corresponding row becomes zero, and then we exploit the \textit{no cone-object} skip to compute the gaze heatmap. The \textit{object score} matrix $\Sigma$ is used by \gazedecoder as an additive bias in the attention module. The rationale behind the score matrix $\Sigma$ is to exploit the strong prior coming from the gaze vector and constrain the network to focus on relevant objects in the scene.

\begin{table*}[!th]
\begin{center}
\resizebox{1.9\columnwidth}{!}{%
\begin{tabular}{lcccccccc}
\hline
\multicolumn{1}{c}{\multirow{3}{*}{Method}}                & \multirow{3}{*}{Modalities} & \multirow{3}{*}{\begin{tabular}[c]{@{}c@{}}Multiperson\\ Gaze\end{tabular}} & \multicolumn{3}{c}{\begin{tabular}[c]{@{}c@{}}GazeFollow \cite{Recasens2015}\\  \end{tabular}}             & \multicolumn{3}{c}{\begin{tabular}[c]{@{}c@{}}VideoAttentionTarget \cite{chong2020detecting}\\ \end{tabular} } \\ \cline{4-9} 
\multicolumn{1}{c}{}                                       &                             &                                                                             & \multirow{2}{*}{}              & \multicolumn{2}{c}{Distance $\downarrow$}                                                   & \multicolumn{2}{c}{\textit{In frame}}                                                          & \textit{Out of frame}                         \\ \cline{5-9} 
\multicolumn{1}{c}{}                                       &                             &                                                                             &  AUC $\uparrow$                                            & Avg.                                         & Min.                                         & AUC $\uparrow$                                 & Dist. $\downarrow$                            & AP $\uparrow$                                 \\
\multicolumn{1}{c}{}                                       &                             &                                                                             &                                        \textit{Head} \textit{Real}$^\dagger$        & \textit{Head} \textit{Real}$^\dagger$                                        & \textit{Head} \textit{Real}$^\dagger$                                         & \textit{Head} \textit{Real}$^\dagger$                                 & \textit{Head} \textit{Real}$^\dagger$                            & \textit{Head} \textit{Real}$^\dagger$                               \\ 
\multicolumn{1}{c}{}                                       &                             &                                                                            &       \hspace{0.2cm}\textit{GT}\hspace{1cm}                                        & \hspace{0.2cm}\textit{GT}\hspace{1cm}                                          & \hspace{0.2cm}\textit{GT}\hspace{1cm}                                          & \hspace{0.2cm}\textit{GT}\hspace{1cm}                                 & \hspace{0.2cm}\textit{GT}\hspace{1cm}                             & \hspace{0.2cm}\textit{GT}\hspace{1cm}                                \\ \hline
Random                                                     &                             &                                                                             & 0.504\hspace{0.2cm}0.391                                        & 0.484\hspace{0.2cm}0.533                                        & 0.391\hspace{0.2cm}0.487                                        & 0.505\hspace{0.2cm}0.247                                         & 0.458\hspace{0.2cm}0.592                                        & 0.621\hspace{0.2cm}0.349                                         \\
Center                                                     &                             &                                                                             & 0.633\hspace{0.2cm}0.446                                        & 0.313\hspace{0.2cm}0.495                                        & 0.230\hspace{0.2cm}0.371                                        & -\hspace{0.7cm} -                                               & -\hspace{0.7cm} -                                             & -\hspace{0.7cm} -                                              \\
Fixed bias                                                 &                             &                                                                             & -\hspace{0.7cm} -                                            & -\hspace{0.7cm} -                                           & -\hspace{0.7cm} -                                             & 0.728\hspace{0.5cm}-\hspace{0.3cm}                                          & 0.326\hspace{0.5cm}-\hspace{0.3cm}                                        & 0.624\hspace{0.5cm}-\hspace{0.3cm}                                         \\
Recasens et al. \cite{Recasens2015}       & R                           &           \xmark                                                                  & 0.878\hspace{0.2cm}0.804                                        & 0.190\hspace{0.2cm}0.233                                        & 0.113\hspace{0.2cm}0.124                                        & -\hspace{0.7cm} -                                               & -\hspace{0.7cm} -                                              & -\hspace{0.7cm} -                                             \\
Chong et al. \cite{Chong_2018_ECCV}     & R                           &     \xmark                                                                        & 0.896\hspace{0.2cm}0.807                                        & 0.187\hspace{0.2cm}0.207                                        & 0.112\hspace{0.2cm}0.120                                        & 0.830\hspace{0.2cm}0.791                                        & 0.193\hspace{0.2cm}0.214                                        & 0.705\hspace{0.2cm}0.651                                       \\
Lian et al. \cite{lian2018believe}        & R                           &    \xmark                                                                         & 0.906\hspace{0.2cm}0.881                                        & 0.145\hspace{0.2cm}0.153                                        & 0.081\hspace{0.2cm}0.087                                        & 0.837\hspace{0.2cm}0.784                                          & 0.165\hspace{0.2cm}0.172                                         & -\hspace{0.7cm} -                                             \\
Chong et al. \cite{chong2020detecting}    & R + T                       &      \xmark                                                                       & 0.921\hspace{0.2cm}0.902                                        & 0.137\hspace{0.2cm}0.142                                        & 0.077\hspace{0.2cm}0.082                                        & 0.860\hspace{0.2cm}0.812                                         & 0.134\hspace{0.2cm}0.146                                        & 0.853\hspace{0.2cm}0.849                                       \\
Fang et al. \cite{fang2021dual}           & R + D                       &      \xmark                                                                       & 0.922\hspace{0.5cm}-\hspace{0.3cm}                                        & 0,124\hspace{0.5cm}-\hspace{0.3cm}                                      & 0.067\hspace{0.5cm}-\hspace{0.3cm}                                       & 0.905\hspace{0.5cm}-\hspace{0.3cm}                                          & 0.108\hspace{0.5cm}-\hspace{0.3cm}                                        & 0.896\hspace{0.5cm}-\hspace{0.3cm}                                        \\
Bao et al. \cite{bao2022escnet}           & R + D + P                           &  \xmark                                                                           & 0.928\hspace{0.5cm}-\hspace{0.3cm}                                        & 0.122\hspace{0.5cm}-\hspace{0.3cm}                                        & -\hspace{0.7cm} -                                            & 0.885\hspace{0.5cm}-\hspace{0.3cm}                                          & 0.120\hspace{0.5cm}-\hspace{0.3cm}                                         & 0.869\hspace{0.5cm}-\hspace{0.3cm}                                         \\
Jin et al. \cite{jin2022depth}            & R + D                       &  \xmark                                                                           & 0.920\hspace{0.5cm}-\hspace{0.3cm}                                       & 0.118\hspace{0.5cm}-\hspace{0.3cm}                                       & 0.063\hspace{0.5cm}-\hspace{0.3cm}                                        & 0.900\hspace{0.5cm}-\hspace{0.3cm}                                          & 0.104\hspace{0.5cm}-\hspace{0.3cm}  & 0.895\hspace{0.5cm}-\hspace{0.3cm}                                         \\
Tonini et al. \cite{tonini2022multimodal} & R + D                       &   \xmark                                                     & 0.927\hspace{0.2cm}0.894                                        & 0.141\hspace{0.2cm}0.165                                        & -\hspace{0.7cm} -                                            & 0.940\hspace{0.2cm}0.894   & 0.129\hspace{0.2cm}0.182                                         & -\hspace{0.7cm} -                                             \\
Qiaomu et al. \cite{miao2023patch}        & R + D + T                   &   \xmark                                                                          & 0.934\hspace{0.5cm}-\hspace{0.3cm} & 0.123\hspace{0.5cm}-\hspace{0.3cm} & 0.065\hspace{0.5cm}-\hspace{0.3cm} & 0.917\hspace{0.5cm}-\hspace{0.3cm}                                          & 0.109\hspace{0.5cm}-\hspace{0.3cm}                                         & 0.908\hspace{0.5cm}-\hspace{0.3cm}  \\ 

Tu et al. \cite{tu2022end}                & R                         & \cmark                                                       & \hspace{0.3cm}-\hspace{0.5cm}0.917                                        & \hspace{0.3cm}-\hspace{0.5cm}0.133                                        & \hspace{0.3cm}-\hspace{0.5cm}0.069                                        & \hspace{0.3cm}-\hspace{0.5cm}0.904                                          & \hspace{0.3cm}-\hspace{0.5cm}0.126                                         & \hspace{0.3cm}-\hspace{0.5cm}0.854                                         \\
Tu et al. \cite{tu2022end}$^\star$        & R                         & \cmark                                                       & \hspace{0.3cm}-\hspace{0.5cm}0.915                                        & \hspace{0.3cm}-\hspace{0.5cm}0.104                                        & \hspace{0.3cm}-\hspace{0.5cm}0.055                                        & \hspace{0.3cm}-\hspace{0.5cm}0.891                                          & \hspace{0.3cm}-\hspace{0.5cm}0.229                                         & -\hspace{0.7cm} 0.809                                              \\ \hline

\rowcolor[HTML]{caf0f8} Our method                                                 & R                           & \cmark                                                       & \hspace{0.3cm}-\hspace{0.5cm}\textbf{0.922}  & \hspace{0.3cm}-\hspace{0.5cm}0.072  & \hspace{0.3cm}-\hspace{0.5cm}0.033  & \hspace{0.3cm}-\hspace{0.5cm}0.923    & \hspace{0.3cm}-\hspace{0.5cm}\textbf{0.102}   & \hspace{0.3cm}-\hspace{0.5cm}\textbf{0.944}   \\
\rowcolor[HTML]{caf0f8} Our method                                                 & R + D                       & \cmark                                                       & \hspace{0.3cm}-\hspace{0.5cm}\textbf{0.922}                                            & \hspace{0.3cm}-\hspace{0.5cm}\textbf{0.069}                                            & \hspace{0.3cm}-\hspace{0.5cm}\textbf{0.029}                                            & \hspace{0.3cm}-\hspace{0.5cm}\textbf{0.933}                                              & \hspace{0.3cm}-\hspace{0.5cm}0.104                                             & \hspace{0.3cm}-\hspace{0.5cm}0.934                                       \\ \hline
\end{tabular}
}
\caption{Evaluation on the GazeFollow \cite{Recasens2015} and VideoAttentionTarget \cite{chong2020detecting} datasets. \textit{Head GT} refers to using carefully labeled ground-truth head crops and head locations in training and testing. \textit{Real} indicated with $^\dagger$ is the implementation of \cite{tu2022end}, which applies an additional SOTA head detection network to predict the head location for real-world applications. We produce only \cite{tonini2022multimodal}'s \textit{Real} results (see text for details).
$^\star$ indicates our implementation. $R$, $D$, $T$, and $P$ stand for RGB, depth, temporal processing, and 2D-pose, respectively. Refer to Supp. Mat. for Angular Error results.}
\label{table:sota}
\vspace{-0.7cm}
\end{center}
\end{table*}

\subsection{Gaze Object Transformer}
\vspace{-0.1cm}
Although the information from the predicted gaze vector, cone, and $\Sigma$ provides important knowledge for the task at hand, accurately predicting the gaze direction is fundamentally a hard problem since the precise angle and magnitude of the vector are highly sensitive and might even introduce noise for the training procedure (see Sec.~\ref{sec:ablation} for empirical justification). Eventually, an accurate vector estimation requires considering eye position and sight. However, such elements potentially introduce the need to use auxiliary networks, increasing the computational complexity of the overall architecture. Instead, our proposal is much simpler but effective as it does not use the gaze vector to predict the final heatmap, but we further process the output of $\mathcal{D}$ with the aid of the \textit{object score} matrix $\Sigma$ in \gazedecoder, which follows the same design principle as the decoder $\mathcal{D}$.

First, a stack of MHSA and FFN layers encodes a set of learnable embeddings $\mathbf{e}_g \in \mathbb{R}^{N \times C_b^{'}}$, where $N$ is the number of predicted objects. Unlike the object detector transformer's encoder, the multi-head self-attention includes an additive bias, \ie our \textit{object score} matrix $\Sigma$. Therefore, the new attention is defined as:
\vspace{-0.2cm}
\begin{equation}
\small{\text{BiasedAttention}(Q,K,V) = \text{softmax}\Big(\frac{QK^T + \Sigma}{\sqrt{d_k}}\Big)V }
\vspace{-0.1cm}
\end{equation}
Additionally, we mask the learnable embeddings corresponding to objects not classified as heads.
The masked features of the self-attention of \gazedecoder are the inputs to the cross-attention module.
Likewise self-attention, the cross-attention module exploits the \textit{object score} matrix as additive bias and performs binary masking on heads for $Q$ and other objects for $K$ and $V$. We also exclude the objects with low confidence prediction or that are classified as \textit{no-object} ($\emptyset$) (see Sec.~\ref{sec:impDet} for details).

The output features of the cross-attention form the input to the \textit{heatmap} MLP to predict the gaze heatmap for each head.
However, since we cannot assume that an object is always present, a second MLP (\textit{heatmap no-object} in Fig.~\ref{fig:architecture}) predicts the heatmap from head features only when no object is inside the visual cone.
The outputs of \textit{heatmap} MLP and \textit{heatmap no-object} MLP are fed to \textit{a gated operator} that selects the heatmap based on the presence (or absence) of objects in the cone of each person. Finally, an additional \textit{watch outside} MLP, only for head objects, predicts $\mathbf{p}_{out}$, the probability that the given head gaze lies outside the frame.

\subsection{Training objective}
\vspace{-0.2cm}
As we perform multiple tasks simultaneously (e.g., object localization and classification, gaze vector regression, and gaze heatmap regression), our training objective is defined as a \textit{\textbf{weighted sum}} of all tasks.

We supervise the object localization with the weighted difference of $\mathcal{L}_1$ distance and Generalized Intersection over Union (GIoU) \cite{Rezatofighi_2018_CVPR} of the target box $box$ and predicted box $box_p$, respectively, formalized as $\mathcal{L}_{box} = \lambda_{l1} \| box - box_p \| - \lambda_{giou} GIoU(box, box_p)$.
Object classification loss $\mathcal{L}_{cls}$ is the cross-entropy between the ground truth label and the post-softmax distribution of the predicted class.

The gaze-related tasks involve the use of three losses: (a) gaze vector loss, (b) gaze heatmap loss, and (c) gaze watch-outside loss.
The gaze vector loss is formulated as the $\mathcal{L}_2$ loss between elements of the predicted and target vector such that $ \mathcal{L}_{vec} = \| \mathbf{v}_g - \mathbf{v}_p \|_2 $, with \gazevector being the ground truth gaze vector and $\mathbf{v}_p$ the one predicted by our method. 
The watch-outside loss is a binary cross-entropy loss $\mathcal{L}_{out} = -\big[out\log(\mathbf{p}_{out}) + (1-out)\log(1-\mathbf{p}_{out})\big]$, where $out$ is the ground truth binary annotation of whether the person is watching outside and $\mathbf{p}_{out}$ is the predicted value.
Lastly, the gaze heatmap loss is an $\mathcal{L}_2$ loss between target and predicted heatmap: $\mathcal{L}_{heat} = \lambda_{heat} \| \mathbf{H} - \mathbf{H}_p\|_2$.

\vspace{-0.2cm}

\section{Experiments}
\label{sec:exp}
\vspace{-0.1cm}
\subsection{Datasets and Evaluation metrics}
\vspace{-0.1cm}
\paragraph{Datasets.} Our model is trained and tested on both GazeFollow \cite{Recasens2015} and VideoAttentionTarget \cite{chong2020detecting} datasets. \textbf{GazeFollow} \cite{Recasens2015} is a large-scale \emph{image} dataset containing over 122K images in total with more than 130K people. The test images include gaze and head location annotations performed by up to 10 people for a single person in the scene while the training set contains only one annotator's judgment indicating gaze and head locations. \textbf{VideoAttentionTarget} \cite{chong2020detecting} is composed of YouTube \emph{video} clips, each has a length of up to 80 seconds. It includes 109574 in-frame and 54967 out-of-frame gaze annotations together with the head locations. Both the training and test sets contain one gaze annotation per person. Given that we do not use the \emph{temporal} information in our model, we randomly select one image for every 5 consecutive frames, allowing us to avoid overfitting. This setup is the same with SOTA~\cite{bao2022escnet,fang2021dual,jin2022depth,tonini2022multimodal,tu2022end}.  \\
\vspace{-0.3cm}

\noindent \textbf{Evaluation Metrics.} We evaluate the performance of the proposed method in terms of \textbf{gaze target detection} and \textbf{object class detection and localization}. For the former task, we use all standard metrics~\cite{Chong_2018_ECCV,chong2020detecting} described as follows. \textbf{AUC} assesses the confidence of the predicted gaze heatmap \textit{w.r.t.} the gaze ground-truth. \textbf{Distance} (Dist.) is the $\mathcal{L}_2$ between the ground-truth gaze point and the predicted gaze location, which is the point with the maximum confidence on the gaze heatmap. \textbf{Angular Error} (Ang. Err.) is the angle between predicted and ground-truth gaze vector. In GazeFollow, it is a standard to declare both the minimum and average distances. \textbf{I/O gaze AP} is the average precision used to evaluate the \emph{out-of-frame} probability of the gaze in VideoAttentionTarget. We use the standard metric \textbf{Mean Average Precision (mAP)} for object class detection and localization. In that case, a prediction is correct if the class label of the predicted bounding box and the ground truth bounding box are the same and the Intersection over Union ($IoU$) between them is greater than a $threshold$ value.
\vspace{-0.1cm}
\subsection{Implementation details}
\label{sec:impDet}
\vspace{-0.1cm}
$\mathcal{B}$ is a ResNet-50 pretrained on ImageNet~\cite{imagenet_cvpr09} and the Object Detector Transformer follows the DETR~\cite{DETR} architecture. We train all our components (Object Detector Transformer, \gazedecoder, \textit{Gaze Cone Predictor}, and MLPs) with Adam optimizer and a learning rate of $1 \times 10^{-4}$ for 80 epochs, then we drop the learning rate by 10 times and train for 20 epochs. Differently, $\mathcal{B}$ has a learning rate 10 times smaller, i.e. $1 \times 10^{-5}$.
Furthermore, we perform matching between predictions and ground-truth samples as described in \cite{tu2022end}.
The FoV angle of the cone predictor is set to $120\degree$, corresponding to the binocular FoV of humans~\cite{howard1995binocular}.
\gazedecoder keeps only queries of objects classified as heads and with confidence above $0.5$. Conversely, the keys and values are those of objects (heads included) with the confidence above $0.5$, which are not classified as \textit{no-object}.
The final loss is the weighted sum of the defined objectives. We set $\lambda_{gious}=2.5$ and $\lambda_{heat}=2$. The other losses are summed up without any weighting.
We use a SoTA monocular depth estimator \cite{ranftl2020towards} to obtain depth maps corresponding to each scene image.
Note that we use depth information only for gaze cone building without learning additional depth features.
More details are available in Supp. Mat.

\subsection{Comparison with State-of-the-Art}
\label{sec:compSOTA}
\vspace{-0.1cm}
The gaze target detection performance of our method is compared with the SOTA in Table~\ref{table:sota}. Recalling that the cropped head images and the head locations are required for traditional methods (i.e., SOTA except~\cite{tu2022end}) and these methods are evaluated when the ground-truth head locations are granted (referred to as ``Head GT''), we proceed with the evaluation procedure of~\cite{tu2022end}, summarized as follows. Tu et al.~\cite{tu2022end} employ additional head detectors to automatically obtain the heads position, which is given to the traditional models, providing their real-world application performance. We inherit the corresponding results from~\cite{tu2022end} and refer to them as ``Real''. For the methods whose ``Real'' results are not provided by~\cite{tu2022end}, we obtain the results using RetinaFace~\cite{Deng_2020_CVPR} to detect heads position. However, we are able to perform this only for the method whose code is publicly available:~\cite{tonini2022multimodal}. 

As we can see from the results, our method only with RGB data outperforms existing SOTA on all datasets for all metrics. Such a performance is important to emphasize since several SOTA perform relatively poorly even though they use multi-modalities \cite{fang2021dual,jin2022depth} or temporal data \cite{chong2020detecting}. Particularly, for VideoAttentionTarget \cite{chong2020detecting} dataset, our method achieves better scores compared to many complex methods relying on several pretrained task-specific backbones (e.g., 2D-pose estimation) \cite{bao2022escnet} or leveraging the temporal dimensionality of the data \cite{miao2023patch} while both utilize RGB and depth maps.
Our better performance \textit{w.r.t.} Transformer-based \cite{tu2022end} is also conspicuous.
Furthermore, when both RGB and depth are taken into account, our method performance slightly improves on average. Recalling that we use depth information only during gaze cone production without requiring additional (pretrained) CNN to learn depth features as in \cite{jin2022depth,tonini2022multimodal} or needing to detect the eyes as in \cite{fang2021dual}, the corresponding results are momentous. Particularly, our minimum and average distance and mAP results are always the best whether or not others were evaluated within ``Head GT'' or ``Real'' settings. This shows that the proposed method is notably good at predicting if the gaze is located inside or outside the frame, the gaze heatmaps, and eventually, a single pixel gaze point that our model predicts per person is much closer to the ground truth-gaze point.

\subsection{Ablation Study}
\label{sec:ablation}
\vspace{-0.1cm}
The ablation study is performed on both GazeFollow~\cite{Recasens2015} and VideoAttentionTarget~\cite{chong2020detecting} datasets, whose results can be found in Table~\ref{tab:ablation_complete}. \\
\noindent\textbf{Gaze Object Transformer.}
If we do not use \gazedecoder, it is still possible to predict the 
gaze heatmap using the features of $\mathcal{D}$. As seen from the results (first row of the ablations for each dataset), $\mathcal{D}$ features alone are insufficient to reach SOTA results for gaze target detection. Whereas including \gazedecoder boosts the results for all metrics and datasets (second row of the ablations for each dataset). \\
\noindent\textbf{Object Masking.} By definition, Transformer attention attends to every token in a sequence and tries to learn relationships between all elements. In our case, this refers to computing the interaction between every object.
Instead, our design retains only the \textit{queries} to be of those elements recognized as heads and \textit{keys} and \textit{values} to be those of any other object/head. In this way, we obtain an improvement across both datasets for all the metrics (third row of the ablations). Furthermore, we obtain an interpretable attention matrix of interaction between heads and objects. \\
\noindent\textbf{Gaze Cone and No cone-object Skip.} 
Gaze cone building assigns a score into $\Sigma$  inversely proportional to the distance from the gaze vector for the objects inside the cone. This acts similarly to a temperature to skew the softmax operation inside the attention towards the objects more probable to be looked at. As seen, the gaze cone alone might not be sufficient to improve the performance of the method (fourth row of the table). We attribute this to the cases where we cannot find a meaningful object inside the gaze cone, meaning that the $\Sigma$ row corresponding to the face is empty, and attention does not operate on any feature, hindering the performance of the \textit{heatmap} MLP. To solve this, we design a \textit{no cone-object} skip, which allows building a heatmap starting from $\mathcal{D}$ features. In such cases, a gating mechanism allows selecting which heatmap to use depending on the presence of objects in the cone. When we use this mechanism in conjunction with cone building (fifth row of the table, aka \textit{full proposed method}), we obtain the best results consistently across the datasets, proving the effectiveness of focusing on relevant objects in the scene.

\begin{table}[!t]
\centering
\resizebox{\linewidth}{!}{%
\begin{tabular}{@{}lcccccc@{}}
\toprule
\multirow{2}{*}{\gazedecoder} &
  \multirow{2}{*}{OM} &
  \multirow{2}{*}{GC} &
  \multirow{2}{*}{NOCS} &
  \multicolumn{3}{c}{GazeFollow \cite{Recasens2015}}  \\ \cmidrule(l){5-7} 
       &        &        &        & \multicolumn{1}{c}{AUC $\uparrow$}   & Avg. dist. $\downarrow$  & Min. dist. $\downarrow$  \\ \midrule
\xmark & \xmark & \xmark & \xmark & \multicolumn{1}{c}{0.864} & 0.110 & 0.061 \\
\cmark & \xmark & \xmark & \xmark & \multicolumn{1}{c}{0.918} & 0.075 & 0.038 \\
\cmark & \cmark & \xmark & \xmark & \multicolumn{1}{c}{0.919} & 0.073 & 0.033 \\
\cmark & \cmark & \cmark & \xmark & \multicolumn{1}{c}{0.905} & 0.090  & 0.051 \\
\cmark & \cmark & \cmark & \cmark & \multicolumn{1}{c}{\textbf{0.922}} & \textbf{0.072}  & \textbf{0.033} \\ \bottomrule
\multirow{2}{*}{\gazedecoder} &
  \multirow{2}{*}{OM} &
  \multirow{2}{*}{GC} &
  \multirow{2}{*}{NOCS}  &
  \multicolumn{3}{c}{VideoAttentionTarget \cite{chong2020detecting}} \\ \cmidrule(l){5-7} 
      &        &        &        &  \multicolumn{1}{c}{AUC~$\uparrow$} & Dist.~$\downarrow$ & AP~$\uparrow$ \\ \midrule
\xmark & \xmark & \xmark & \xmark &  \multicolumn{1}{c}{0.811} & 0.271  & 0.77  \\
\cmark & \xmark & \xmark & \xmark &  \multicolumn{1}{c}{0.902} & 0.125  & 0.92 \\
\cmark & \cmark & \xmark & \xmark &  \multicolumn{1}{c}{0.907} & 0.112  & \textbf{0.94} \\
\cmark & \cmark & \cmark & \xmark &  \multicolumn{1}{c}{0.909} & 0.154  & \textbf{0.94} \\
\cmark & \cmark & \cmark & \cmark &  \multicolumn{1}{c}{\textbf{0.923}} & \textbf{0.101}  & \textbf{0.94} \\\bottomrule
\end{tabular}
}
\vspace{-0.2cm}
\caption{Ablation study on GazeFollow~\cite{Recasens2015} and VideoAttentionTarget~\cite{chong2020detecting}. $OM$, $GC$, $NCOS$ stand for object masking, gaze cone, and \textit{no cone-obj} skip, respectively.}
\label{tab:ablation_complete}
\vspace{-1em}
\end{table}

\vspace{-0.25cm}
\subsection{Gazed-object class detection and localization}
\label{sec:gazedObjectEval}
\vspace{-0.1cm}

\begin{table} [!htb]
\centering
\resizebox{\linewidth}{!}{
\begin{tabular}{@{}llrrr@{}}
\toprule
Method                                    & \# params.~$\downarrow$ & \multicolumn{1}{l}{AP~$\uparrow$} & \multicolumn{1}{l}{AP$_{50}$$~\uparrow$} & \multicolumn{1}{l}{AP$_{75}$~$\uparrow$} \\ \midrule
Tu et al. \cite{tu2022end}         &   43M        & 0.01                  & 0.03                         & 0.01                         \\
Tu et al. \cite{tu2022end} + DETR \cite{DETR} &  84M         & 0.04                  & 0.12                         & 0.02                         \\
Ours + DETR \cite{DETR}                    &    97M       & 0.03                  & 0.10                         & 0.01                         \\
Ours                                       &    \textbf{54M}       & \textbf{0.14}         & \textbf{0.22}                & \textbf{0.15}                \\ \bottomrule
\end{tabular}%
}
\vspace{-0.2cm}
\caption{Gazed-object classification and localization performance. The computational complexity is reported in terms of parameters.}
\label{table:objectdetect}
\vspace{-0.5cm}
\end{table}

This section reports the evaluations regarding gazed-object class detection and localization performance.
Our method is notably different from using an auxiliary object detector accompanying a gaze target detection model. Still, in order to empirically highlight the difference, we combine the model of Tu et al.~\cite{tu2022end} 
with DETR~\cite{DETR} pretrained on COCO~\cite{lin2014microsoft}. To this end, given a produced gaze heatmap of~\cite{tu2022end}, we use the bounding box proposal of DETR which contains the highest value of the heatmap (notice that this is in line with ground-truth gaze heatmap construction \cite{chong2020detecting}). 
Similarly, we also combined the results of DETR with our model's gaze heatmap predictions.
Moreover, we include \cite{tu2022end} in the comparisons by determining a bounding box that surrounds the gaze heatmaps of \cite{tu2022end}. In that case, $AP$ was calculated only for object locations, discarding the object class prediction.
The corresponding results given in Table~\ref{table:objectdetect} were performed using the COCO-subset of the GazeFollow dataset \cite{Recasens2015} providing the ground-truth object class and location information. Notice that our model was not particularly trained on COCO ground-truth object classes and locations but was trained on the full set of the GazeFollow and its \emph{gaze} annotations. Instead, DETR was trained on the full COCO dataset \cite{lin2014microsoft}. That setting should rather be advantageous for DETR since it is aware of all object classes. Overall, the results show the relative effectiveness of our model for gazed-object prediction while it is also the most efficient in terms of the number of parameters.

\begingroup
\begin{table}[htb]
\vspace{-0.5em}
\centering
\captionsetup{type=table}
\resizebox{1\linewidth}{!}{%
\begin{tabular}{@{}lrrrrrrrrrr@{}}
\toprule
 & 10\%  & 20\%   & 30\%   & 40\%   & 50\%   & 60\%   & 70\%   & 80\%   & 90\%   & 100\%     \\ \hline
Ours 2D       & 0.980 & 0.977 & 0.973 & 0.970 & 0.964 & 0.961 & 0.958 & 0.954 & 0.943 & 0.922 \\
Ours 3D       & 0.980 & 0.977 & 0.972 & 0.967 & 0.961 & 0.957 & 0.953 & 0.950 & 0.944 & 0.922 \\
\cite{tu2022end} & 0.973 & 0.967 & 0.964 & 0.957 & 0.953 & 0.948 & 0.945 & 0.940 & 0.932 & 0.915 \\
\bottomrule
\end{tabular}%
}
\captionof{table}{AUC between our 2D and 3D method and \cite{tu2022end} w.r.t. increasing variance levels.}
\label{tab:variance}
\vspace{-0.5cm}
\end{table}
\endgroup

\subsection{The effect of variance in gaze annotations}
\vspace{-0.1cm}
We compare the AUC of the proposed method with \cite{tu2022end} considering the multiple annotations that the GazeFollow dataset's test split provides. In some cases, the annotators' consensus is low, as highlighted in \cite{miao2023patch}, which motivated us to evaluate the methods under different levels of variance across the individual gaze annotations. The calculation of the annotation variance and extensive discussions are given in Supp. Mat. The results presented in Table~\ref{tab:variance} show the permanent better performance of our model both in 2D and 3D \textit{w.r.t.} \cite{tu2022end} while, as expected, with lower variance all methods perform better. We speculate that the lower performance of Ours-3D \textit{w.r.t.} Ours-2D can be since the human annotations were collected on 2D images.

\subsection{Qualitative Results}
\label{sec:qualitative}
\begin{figure}[!h]
\centering
\includegraphics[width=0.85\linewidth]{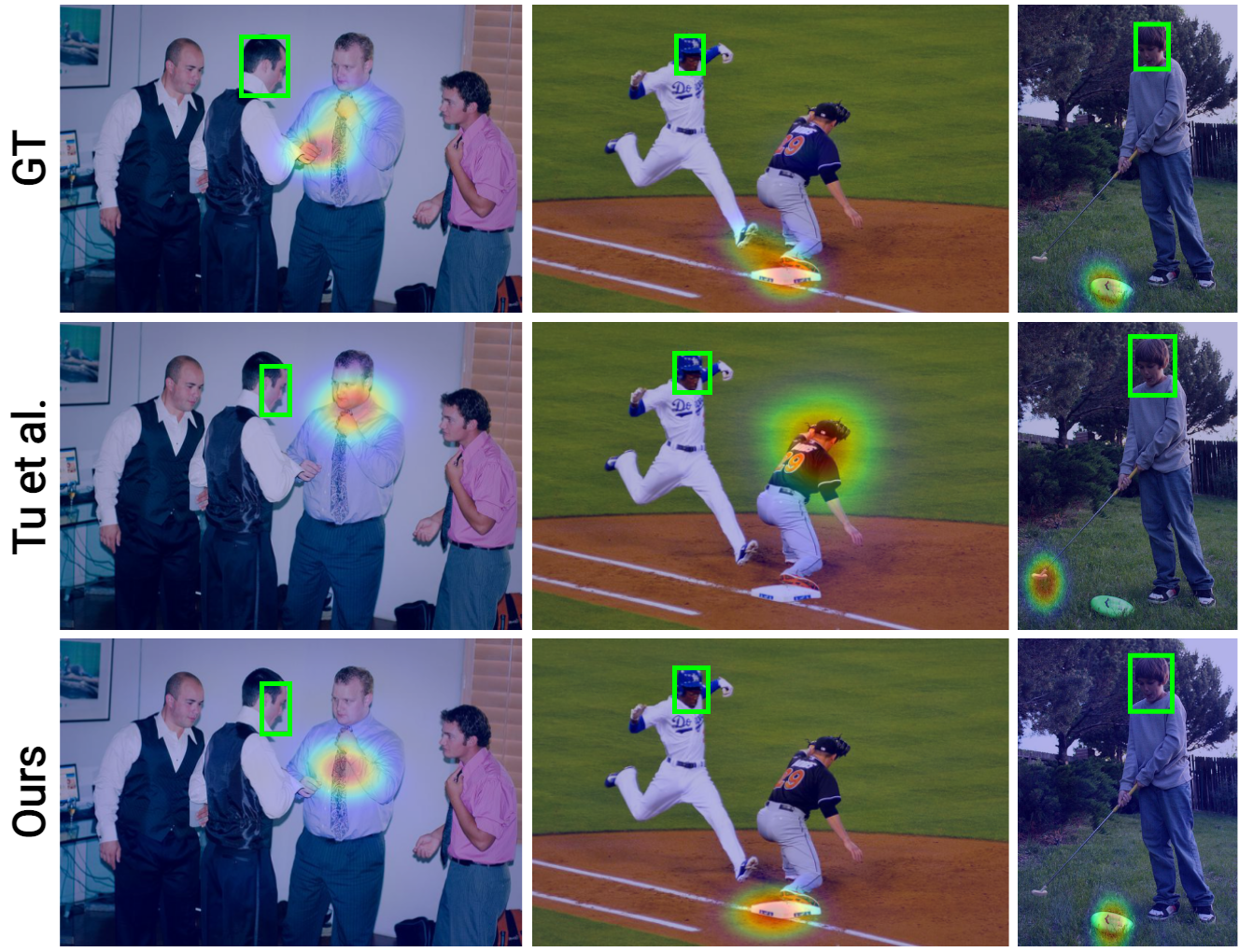}
\vspace{-0.2cm}
\caption{Qualitative results of our method (bottom) and Tu et al. \cite{tu2022end} (middle) \textit{w.r.t.}~the ground-truth (top). For simplicity, we show only one person's gaze.}
\label{fig:qualitatives}
\vspace{-0.5cm}
\end{figure}

\vspace{-0.1cm}
We visualize gaze heatmaps of our method and \cite{tu2022end} in Fig.~\ref{fig:qualitatives} on the GazeFollow dataset. Our predictions are more accurate compared to \cite{tu2022end} in line with the quantitative results. Refer to Supp. Mat. for more qualitative results, including some less accurate performance of the proposed method \textit{w.r.t.} the ground-truth.

\section{Discussion \& Conclusion}
\label{sec:conclusions}
\vspace{-0.1cm}
We have presented a new end-to-end Transformer-based gaze target detector simultaneously predicting the \emph{object class} and the \emph{location} of the gazed-object. The latter is advantageous \textit{w.r.t.} prior art as it improves explainability. 
Extensive experiments validate our approach's better performance for gaze behavior understanding, promising its usefulness in real-world human interaction analysis. \\
\noindent \textbf{Broader Impacts.} 
We target a human-centric task and consequently, our model, in some cases, might need to process human faces. This might result in issues regarding privacy protection, therefore policy review should be considered when using this model in real-world applications. \\
\noindent \textbf{Limitations \& Future Work.} As expected from a Transformer-based model, our network also has slow convergence, requiring long training epochs. Future work will investigate gaze-target detection within the open-set object detection paradigms. 
\vspace{-0.2cm}
\\
\\
\noindent \textbf{Acknowledgments.} This work was supported by the EU H2020 SPRING (No. 871245) and AI4Media (No. 951911) projects, and the MUR PNRR project FAIR - Future AI Research (PE00000013) funded by the NextGenerationEU. It was carried out in the Vision and Learning joint laboratory of FBK and UNITN.

{\small
\bibliographystyle{ieee_fullname}
\bibliography{egbib}
}

\newpage
\appendix
\renewcommand{\thesubsection}{\Alph{subsection}}

\section{The effect of variance in gaze annotations}
\label{sec:variance}

\begin{figure}[ht!]
\begin{center}
\includegraphics[width=\linewidth]{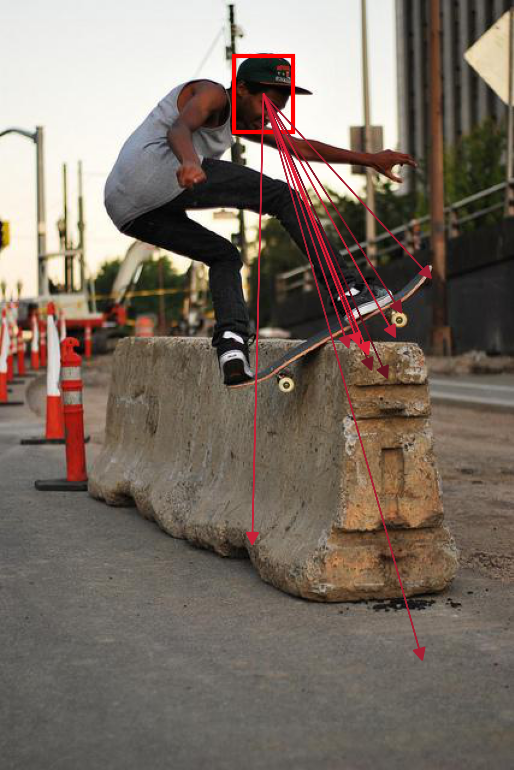}
\end{center}
\caption{An image from the GazeFollow dataset's \cite{Recasens2015} test split which has several annotations for the gaze point with high annotation variance.}
\label{fig:annotations}
\end{figure}

\begin{figure}[ht!]
\begin{center}
\includegraphics[width=\linewidth]{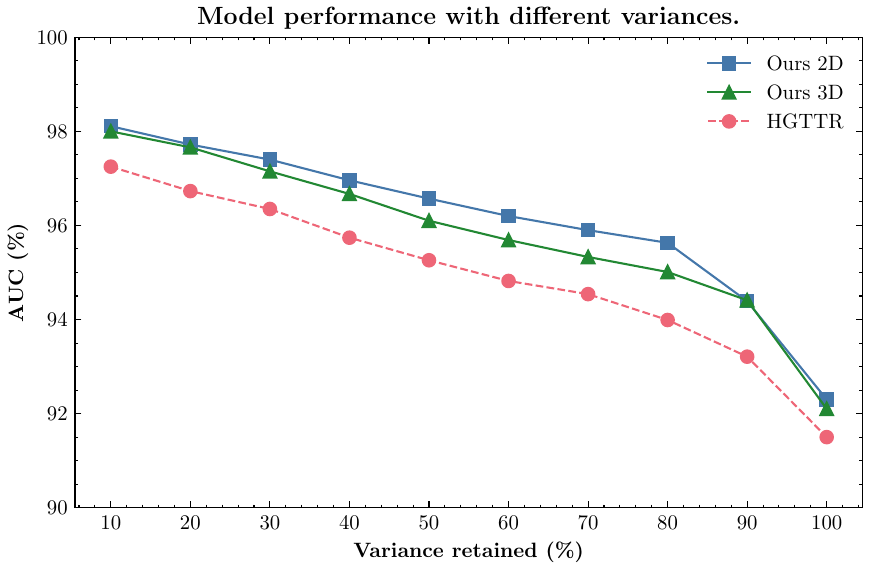}
\end{center}
\vspace{-1.5em}
\caption{Performance of Our model (2D and 3D) and Tu et al.~\cite{tu2022end} \textit{w.r.t.} the variance in the ground-truth annotations of the GazeFollow dataset \cite{Recasens2015}.}
\label{fig:variance}
\vspace{-1.5em}
\end{figure}

As mentioned in the paper, the GazeFollow dataset~\cite{Recasens2015} contains a single gaze point annotation for a single person in a scene in its training split. However, its test splits include several numbers of annotations with respect to a single person's gaze. The number of annotations can be varying up to 10 different gaze points for each person. Such an annotation procedure would not present an issue if all the annotators reached a consensus regarding the gaze point, however, as also shown in \cite{miao2023patch} the test split annotations of that dataset per person can vary remarkably. An example image and the corresponding gaze points demonstrating the variety across the annotations are given in Fig.~\ref{fig:annotations}. 

On the other hand, the standard metric used to make evaluations on this dataset, aka AUC does not consider the (possible) varieties across the annotations. 
For this reason, herein as well as in the main paper, we present an additional evaluation procedure, which considers the multiple annotations that GazeFollow test split provides (see Fig.~\ref{fig:variance}). That evaluation procedure can be described as follows. For each gaze point annotation corresponding to a single person, we compute its distance to the corresponding average gaze annotation point. We record all the distances for the whole test split and compose a distribution from them. Then, given this distribution, we keep the gaze points falling inside a certain threshold (shown as variance retained in the figure), and we opt for the deciles for easiness of computation. For each decile, we compute the AUC and report it in Fig.~\ref{fig:variance}. As seen, our method is extremely effective when there is high annotation consensus, \ie the distance from the average point falls in the first decile (\ie, 0-10\% variance retained in the figure); the performance slightly decreases until the eighth decile (\ie, 70-80\% variance retained in the figure), with the last $20\%$ representing high noise annotations (\ie, 80-100\% variance retained in the figure) where the performance lowers at a faster rate. When we compare our performance against the state-of-the-art method of~\cite{tu2022end}, one can observe a consistently higher performance for all cases both in 2D and 3D versions of our method. We speculate that the lower performance of Ours-3D \textit{w.r.t.} Ours-2D can be since the human annotations were collected on 2D images.

\section{Additional evaluation on GazeFollow \cite{Recasens2015} and VideoAttentionTarget \cite{chong2020detecting}}
Table~\ref{tab:angularerror_results} reports the \textit{Angular Error}~\cite{Recasens2015} (i.e. the angle between predicted and ground-truth gaze vector) results and compare it with SOTA. Our method produces the best results out of all, while Ours (3D) is better than Ours (2D).

\begin{table}[htb]
\begin{center}
\centering
\resizebox{\linewidth}{!}{%
\begin{tabular}{@{}ccc|cccccccc@{}}
\toprule
\multicolumn{1}{l}{} & \multicolumn{1}{l}{Ours (2D)} & \multicolumn{1}{l|}{Ours (3D)}                         & \multicolumn{1}{l}{\cite{tu2022end}$^{\star}$} & \multicolumn{1}{l}{\cite{tonini2022multimodal}$^{\star}$} & \multicolumn{1}{l}{\cite{jin2022depth}} & \multicolumn{1}{l}{\cite{bao2022escnet}} & \multicolumn{1}{l}{{\cite{fang2021dual}}} & \multicolumn{1}{l}{\cite{chong2020detecting}$^{\star}$} & \multicolumn{1}{l}{\cite{lian2018believe}} & \multicolumn{1}{l}{\cite{Recasens2015}} \\ \midrule
Min. $\downarrow$      & 4.0\textdegree                    & \textbf{3.5\textdegree} \textcolor{ForestGreen}{($-12.5\%$)} & 6.6\textdegree                             & 8.1\textdegree                             & ---                          & ---                         & ---                          & 9.1\textdegree                            & 8.8\textdegree                   & ---                          \\
Avg. $\downarrow$      & 7.7\textdegree                    & \textbf{7.2\textdegree} \textcolor{ForestGreen}{($-6.2\%$)}  & 11.0\textdegree                            & 19.5\textdegree                            & 14.8\textdegree                  & 14.6\textdegree                 & 14.9\textdegree                  & 20.5\textdegree                           & 17.6\textdegree                  & 24.0\textdegree                  \\
Max. $\downarrow$      & 20.1\textdegree                   & \textbf{19.3\textdegree} \textcolor{ForestGreen}{($-3.9\%$)} & 22.5\textdegree                            & 37.0\textdegree                            & ---                          & ---                         & ---                          & 37.9\textdegree                           & ---                          & ---                          \\ \bottomrule
\end{tabular}%
}
\caption{Angular error on GazeFollow \cite{Recasens2015} $\star$ means our implementation. Improvements are w.r.t. ``Ours (2D)''.}
\label{tab:angularerror_results}
\end{center}%
\vspace{-1.5em}
\end{table}

\section{Implementation Details}
\label{sec:impDetSup}
We implemented our method in PyTorch and relied on the official code of DETR \cite{DETR} as the backbone.
The heads of DETR \cite{DETR}, \ie the two MLPs for object classification and detection, were replaced by two larger MLPs that allow us to predict the location and classification of objects in the scene including the \textit{heads}. Therefore, the number of classes of objects is adapted to accommodate the \textit{head} class. We used a SOTA object detector, YOLOv8 \cite{Jocher2023}, to pseudo-annotate objects in images that lack object annotations. This has been needed since the used datasets (except the COCO subset of the GazeFollow dataset) do not provide object annotations.
We finetuned the \emph{Object Detector Transformer} using head locations given in the used datasets as well as automatically obtained using an additional head detector, RetinaFace \cite{Deng_2020_CVPR}, and for other objects extracted from YOLOv8. RetinaFace was necessary (but other head detectors can be also adapted as shown in Tu et al. \cite{tu2022end}) as we observed that both Tu et al. \cite{tu2022end}, and our method could not converge without head annotations of all heads in the image.
The depth images were obtained by processing both datasets with a SOTA monocular depth estimation method called MIDAS \cite{ranftl2020towards}.

\section{Qualitative Results}
\label{sec:qualRes}
In this section, we provide additional qualitative results of the gaze heatmaps and the head bounding box of the gaze source (\ie, a person's head) and demonstrate the improved performance of our method \textit{w.r.t.} the current state-of-the-art (SOTA) for both GazeFollow \cite{Recasens2015} and VideoAttentionTarget \cite{chong2020detecting} datasets. Furthermore, we discuss some example cases in which our method has relatively lower performance ($AUC < 70\%$) \textit{w.r.t.} ground-truth as well as Tu et al. \cite{tu2022end}. Lastly, we compare our methods' versions in 2D and 3D and demonstrate the latter's effectiveness in challenging scenarios.

\begin{figure*}[ht!]
\begin{center}
\includegraphics[width=\linewidth,height=0.93\textheight,keepaspectratio]{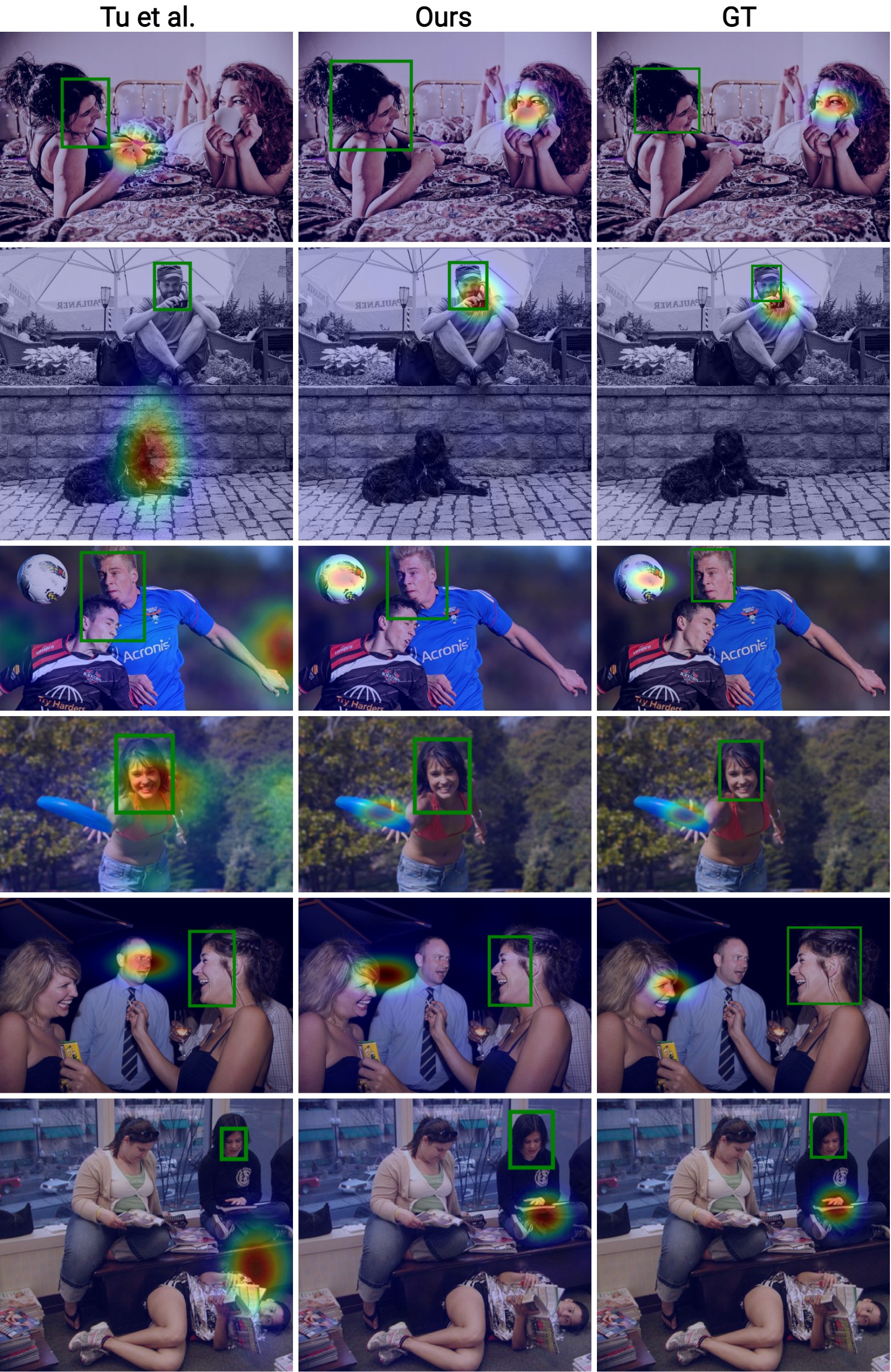}
\end{center}
\vspace{-0.2cm}
   \caption{Qualitative results of our method (center) and Tu et al. \cite{tu2022end} (left) \textit{w.r.t.} the ground-truth annotations of the GazeFollow \cite{Recasens2015} dataset (right). Please note that we only show gaze predictions for the person whose gaze is included in the ground truth. The green boxes show the person-in-interest for the ground truth while they are the detected head for Our and Tu et al. \cite{tu2022end}. Even though our method can detect the gaze of multiple persons in the scene simultaneously, for better visualization, we plot the predicted heatmaps and head locations per person.}
\label{fig:ours_gf_better}
\end{figure*}

\begin{figure*}[ht!]
\begin{center}
\includegraphics[width=\linewidth,height=0.93\textheight,keepaspectratio]{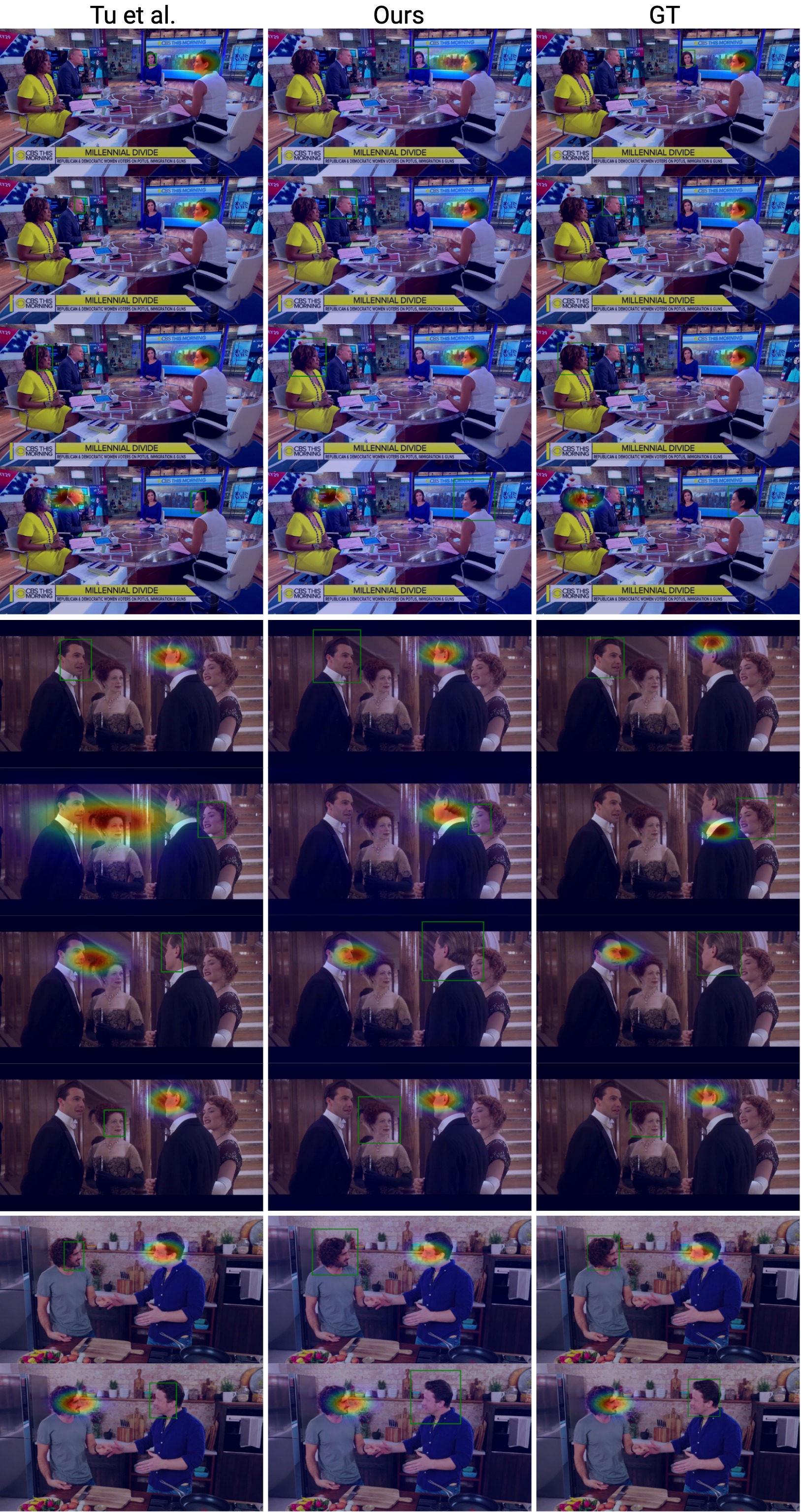}
\end{center}
\vspace{-0.2cm}
   \caption{Qualitative results of our method (center) and Tu et al. \cite{tu2022end} (left) \textit{w.r.t.} the ground-truth annotations of the VideoAttentionTarget \cite{chong2020detecting} dataset (right). Please note that we only show gaze predictions for the person whose gaze is included in the ground truth. The green boxes show the person-in-interest for the ground truth while they are the detected head for Our and Tu et al. \cite{tu2022end}. Even though our method can detect the gaze of multiple persons in the scene simultaneously, for better visualization, we plot a single person's predicted heatmaps and head location.}
\label{fig:ours_videoatt_better}
\end{figure*}

\paragraph{Comparison with SOTA and ground-truth.}
Fig.~\ref{fig:ours_gf_better} and Fig.~\ref{fig:ours_videoatt_better} compare our predictions with respect to the ground truth and the predictions of Tu et al.~\cite{tu2022end} on both datasets, GazeFollow~\cite{Recasens2015} and VideoAttentionTarget~\cite{chong2020detecting}.
As we can see, our model precisely predicts the gaze in many scenes where \cite{tu2022end} is not able to.
More importantly, we can see that predictions of both our method and Tu et al. \cite{tu2022end} are in the field of view of the person whose gaze is to be predicted. However, \cite{tu2022end} favors image regions closer to the gaze source (\ie person's head).

\begin{figure*}[ht!]
\begin{center}
\includegraphics[width=\linewidth,height=0.93\textheight,keepaspectratio]{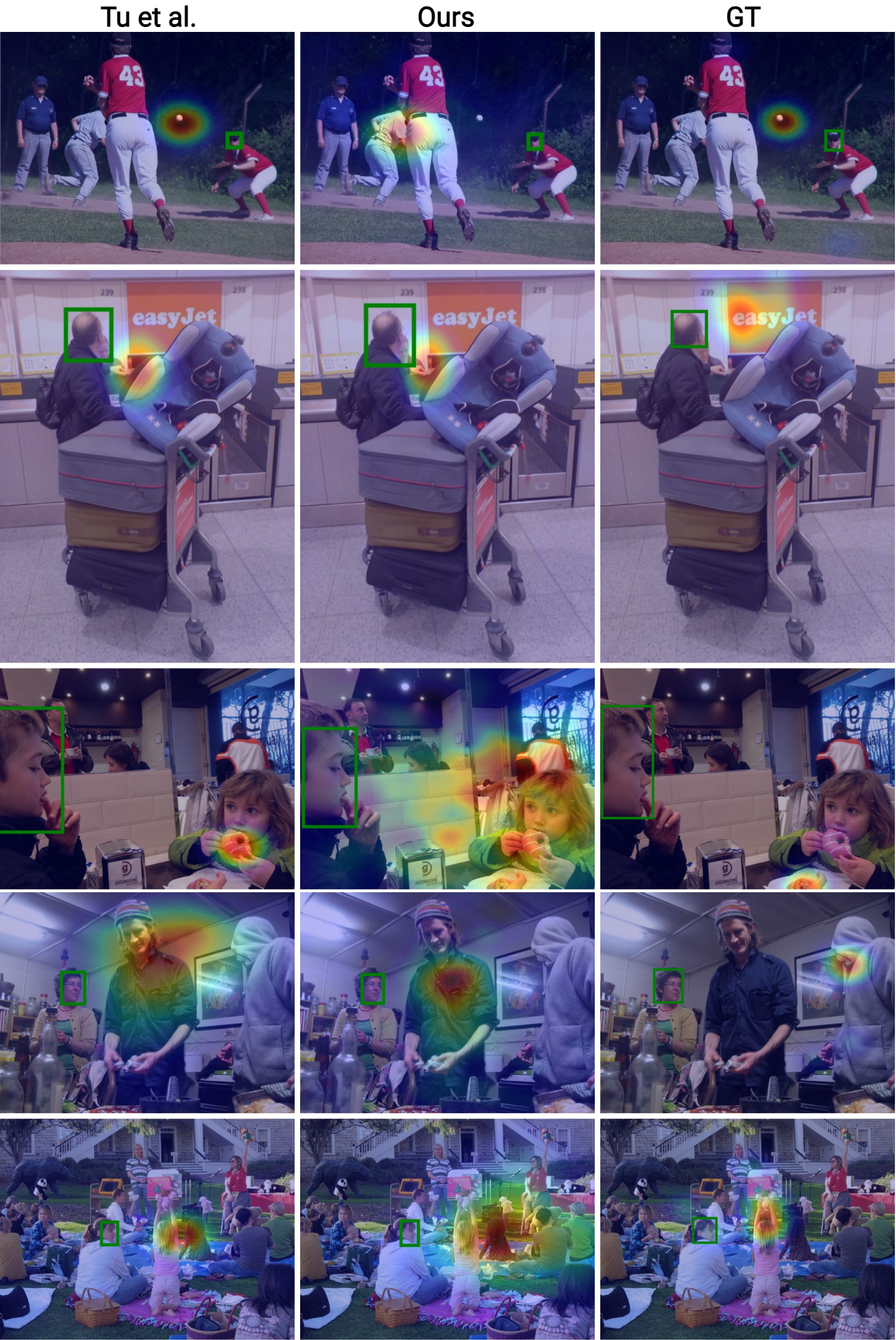}
\end{center}
\vspace{-0.2cm}
   \caption{Qualitative results in which Our method performs relatively lower since the images are highly challenging due to several reasons (see text for details). Our method (center) and Tu et al. \cite{tu2022end} (left) \textit{w.r.t.} the ground-truth annotations of the GazeFollow \cite{Recasens2015} dataset (right). The green boxes show the person-in-interest for the ground truth while they are the detected head for Our and Tu et al. \cite{tu2022end}. Even though our method can detect the gaze of multiple persons in the scene simultaneously, for better visualization, we plot a single person's predicted heatmaps and head location.}
\label{fig:ours_2d_failures}
\end{figure*}

\paragraph{Relatively low-performing predictions.}
Fig.~\ref{fig:ours_2d_failures} presents example images in which our method relatively performs worse. Notice that such images are highly challenging and most of the time also SOTA \cite{tu2022end} underperform, \textit{e.g.} in the second row of Fig.~\ref{fig:ours_2d_failures}, where the head-pose makes it difficult to accurately predict the gaze.
Conversely, when the face is not fully visible, e.g., in the third and fifth row of Fig.~\ref{fig:ours_2d_failures}, we predict scattered heatmaps that cover the gaze point.

\begin{figure*}[ht!]
\begin{center}
\includegraphics[width=\linewidth,height=0.95\textheight,keepaspectratio]{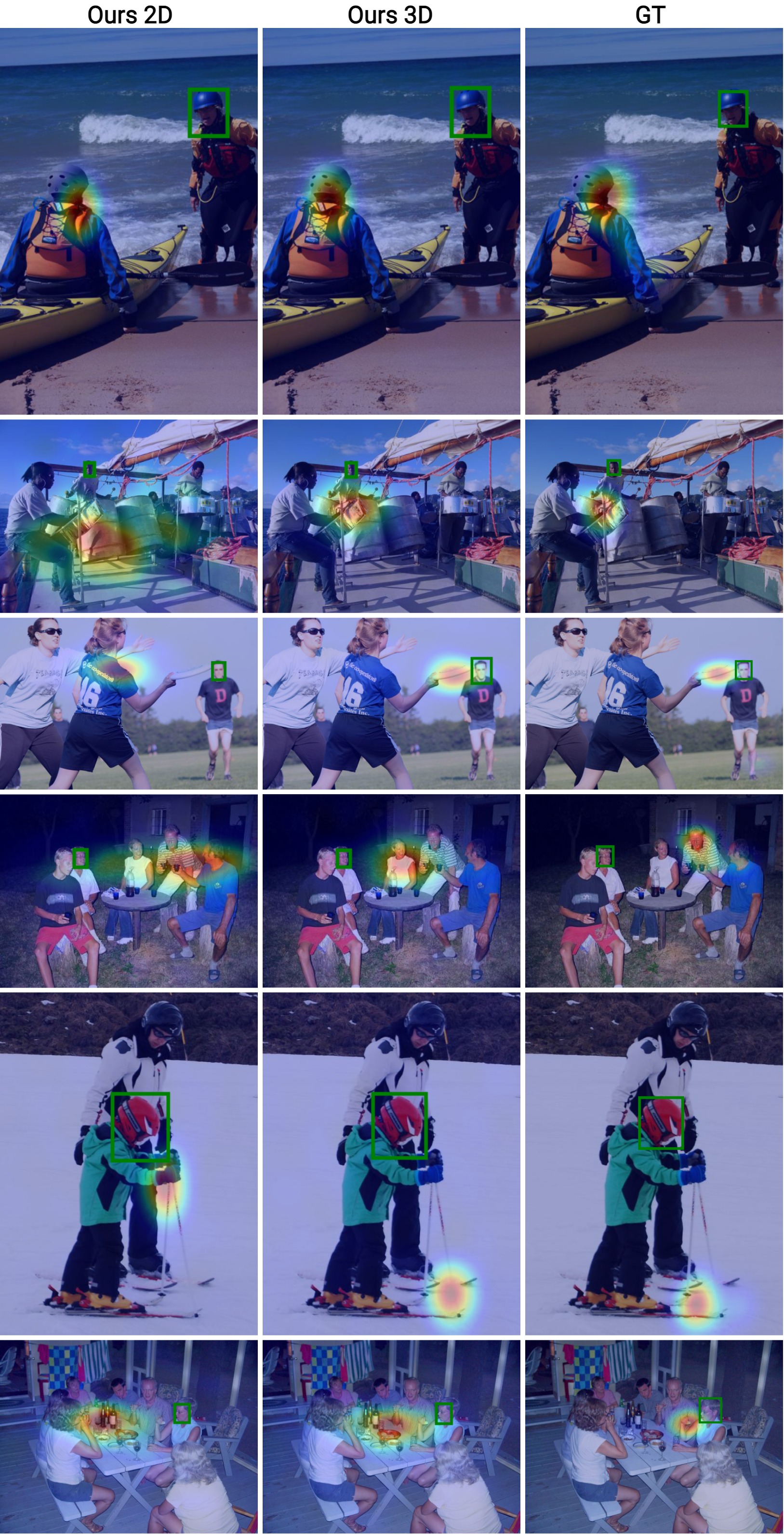}
\end{center}
\vspace{-0.2cm}
   \caption{Qualitative results of our 2D (left) and 3D (center) method \textit{w.r.t.} the ground-truth annotations of the GazeFollow \cite{Recasens2015} dataset (right), showing the importance of 3D-gaze cone building. Even though our method can detect the gaze of multiple persons in the scene simultaneously, for better visualization, we plot a single person's predicted heatmaps and head location.}
\label{fig:ours_2d_3d}
\end{figure*}

\paragraph{The contribution of 3D gaze cone.}
Fig.~\ref{fig:ours_2d_3d} demonstrates the results of our method with the 2D or 3D gaze cone (see main paper for additional details). This comparison aims to highlight the importance of the 3D cone particularly in challenging scenes.
As the quantitative results in the main paper showed, the advantage of the 3D cone is especially visible in terms of the average and minimum distance between the ground-truth gaze point and the point of maximum confidence of the gaze heatmap. The example images in Fig.~\ref{fig:ours_2d_3d} demonstrate that in complex scenes, the 3D gaze vector and the corresponding 3D gaze cone help to decipher which object the person is looking at. Moreover, when the predictions with 2D cone are already high (\textit{e.g.} the first row of Fig.~\ref{fig:ours_2d_3d}), the 3D cone counterpart further consolidates the center of the heatmap towards the object, resulting in better performance.

\end{document}